\documentclass[conference]{IEEEtran}
\IEEEoverridecommandlockouts
\usepackage{cite}
\usepackage{amsmath, amsthm, amssymb, amsfonts}
\usepackage{algorithmic}
\usepackage{graphicx}
\usepackage{mathtools}
\usepackage{textcomp}
\usepackage{xcolor}
\usepackage{hyperref}

\usepackage{caption}

\usepackage{pgfplots}
\pgfplotsset{compat=1.18}
\definecolor{logicblue}{RGB}{105,145,215}
\definecolor{logicyellow}{RGB}{238,190,92}
\usepgfplotslibrary{groupplots}
\definecolor{basec}{RGB}{150,150,150}
\definecolor{sftc}{RGB}{76,120,168}
\definecolor{robonec}{RGB}{228,176,52}
\definecolor{robtwoc}{RGB}{214,95,64}
\definecolor{logicc}{RGB}{88,140,98}
\usepackage[table]{xcolor}
\def\BibTeX{{\rm B\kern-.05em{\sc i\kern-.025em b}\kern-.08em
    T\kern-.1667em\lower.7ex\hbox{E}\kern-.125emX}}

\begin{document}

\newtheorem{theorem}{Theorem}[section]
\newtheorem{proposition}[theorem]{Proposition}
\newtheorem{lemma}[theorem]{Lemma}
\newtheorem{corollary}[theorem]{Corollary}

\newtheorem{definition}[theorem]{Definition}
\newtheorem{assumption}[theorem]{Assumption}
\newtheorem{problem}[theorem]{Problem}
\newtheorem{remark}{Remark} 


\newif\ifhideauthors
\hideauthorsfalse   

\newif\iffull
\fullfalse      

\newif\ifhideack
\hideackfalse   


\newcommand{\authorswitch}[2]{%
  \ifhideauthors #2\else #1\fi
}

\newcommand{\fullswitch}[2]{%
  \iffull #1\else #2\fi
}

\newcommand{\ackswitch}[2]{%
  \ifhideack #2\else #1\fi
}

\title{Logic-VLA: A Temporal Logic Conditioned Vision-Language-Action Model}
\ifhideauthors
\author{}
\else
\author{
Celina Shiyu Wang$^{1}$,
Yiqi Zhao$^{1,\dagger}$,
Junjie Ye$^{1}$,
Yue Wang$^{1}$,
Jyotirmoy V. Deshmukh$^{1,\dagger}$\\
$^{1}$Thomas Lord Department of Computer Science, University of Southern California\\
$^{\dagger}$Equal advising
}
\fi

\maketitle

\begin{abstract}
Vision-language-action (VLA) models can follow natural-language (NL) task instructions, but such instructions may not precisely specify safety-critical or spatiotemporal requirements on the resulting behavior. We introduce \emph{Logic-VLA}, a formal-requirement-aware VLA that conditions on Signal Temporal Logic (STL) specifications supplied at inference time. Logic-VLA uses a syntax-graph-based STL encoder pre-trained to capture temporal logic semantics. Policy adaptation proceeds in two stages: STL-conditioned supervised fine-tuning on satisfying demonstrations is followed by trajectory-level preference optimization over matched satisfying–violating rollout pairs using a flow-matching surrogate for Identity Preference Optimization. This formulation improves formal requirement satisfaction while preserving the nominal NL task. We evaluate Logic-VLA in closed-loop quadcopter navigation simulation across randomized photorealistic environments and test generalization to STL formulas unseen during training. Across the evaluation benchmarks, Logic-VLA improves STL satisfaction rate over an STL-blind base policy by $24.8 \text{ to } 40.7$ percentage points (pp) while reducing nominal NL task success by at most $1.8$ pp, showing that a single VLA can adapt its behavior to varying formal requirements without requiring a separate policy for each specification.
\end{abstract}

\begin{IEEEkeywords}
vision-language-action models, signal temporal logic, preference optimization.\vspace{-5pt}
\end{IEEEkeywords}
\begin{figure*}
    \centering    
    \includegraphics[width=0.72\textwidth]{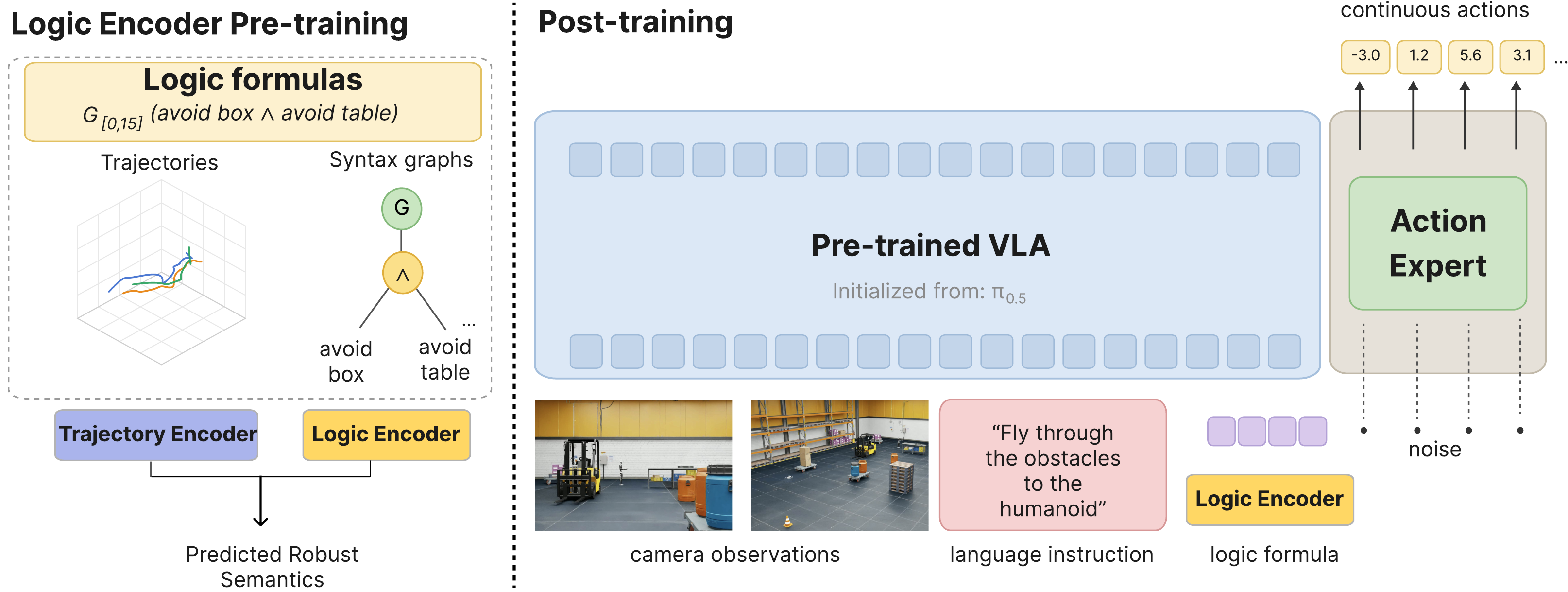}
    \caption{\textbf{Overview of Logic-VLA.} Logic-VLA augments a pre-trained VLA with a structured logic conditioning pathway so that the policy jointly follows a natural-language task and a formal temporal logic requirement. The logic encoder is pre-trained from trajectory-formula pairs using robust semantics supervision (Section \ref{subsec:stl_encoder}). We integrate the encoder with a VLA in a post-training procedure (Section \ref{sec:post-training}) including 1) supervised fine-tuning using logic satisfying demonstrations and 2) trajectory-level preference optimization using matched satisfying-violating trajectory pairs.}
    \label{fig:architecture}
    \vspace{-15pt}
\end{figure*}
\section{Introduction}
A longstanding goal in robotics is to develop agents that flexibly perform diverse tasks in complex environments from intuitive human instructions. Vision-language-action (VLA) models \cite{intelligence2025pi_,zitkovich2023rt} have made significant progress toward this goal by grounding rich visual and language representations in action generation. Given an observation and a natural-language instruction, a VLA produces actions for manipulation, navigation, and other embodied tasks. However, a task instruction alone may not fully specify all desired properties of the resulting behavior. For instance, a robot may need to maintain a prescribed clearance from an object, visit regions in a particular order, complete part of a task within a deadline, or satisfy one condition until another occurs. Such safety-critical, spatiotemporal, and reactive requirements can be difficult to express unambiguously and evaluate precisely using natural-language alone. Signal Temporal Logic (STL) \cite{maler2004monitoring,fainekos2009robustness}, broadly applied in robotics \cite{puranic2021learning}, provides a formal language for specifying such trajectory-level requirements together with robust semantics for evaluating their satisfaction.

This raises a new challenge for VLA models. A deployment-time formal requirement should not replace the natural-language (NL) task, nor should each new requirement require training a separate policy. We seek a single policy whose behavior can adapt to a formal requirement supplied at inference time while retaining the NL task-following capability of a pre-trained VLA. We study \emph{requirement-aware post-training}: given a pre-trained VLA, we adapt the policy to jointly condition on visual observations, the NL task, and an STL specification. Unlike post-training toward a fixed preference or reward, the desired formal requirement is provided as a policy input and may vary across deployments.

To this end, we propose Logic-VLA, a temporal logic-conditioned VLA together with a general post-training framework for incorporating formal trajectory requirements into pre-trained policies. Our instantiation builds on $\pi_{0.5}$ \cite{intelligence2025pi_} and uses a syntax-graph STL encoder built on TeLoGraF \cite{meng2025telograf} to map formal specifications into the VLA conditioning space. We then introduce a two-stage post-training procedure that first imitates STL-satisfying demonstrations and subsequently applies trajectory-level Identity Preference Optimization (IPO) \cite{azar2024general} to distinguish satisfying executions from matched violating alternatives. At inference time, Logic-VLA jointly conditions on the visual observation, the NL task, and a potentially unseen STL specification, allowing a single policy to adapt its behavior to varying formal requirements without retraining for each specification. While our implementation uses $\pi_{0.5}$, the post-training procedure can be applied to other flow-matching VLA policies and provides a general template for requirement-conditioned VLA adaptation. We summarize our contributions:

\begin{itemize}
    \item We introduce Logic-VLA, a $\pi_{0.5}$-based temporal logic conditioned VLA that fuses structured formal requirements with multimodal VLA inputs.
    \item We develop a two-stage requirement-aware post-training framework combining STL-conditioned supervised fine-tuning (SFT) with trajectory-level preference optimization over matched satisfying–violating executions.
    \item We evaluate Logic-VLA on a closed-loop quadcopter navigation benchmark across STL formulas seen and unseen during training, where it improves STL satisfaction over the STL-blind base policy by $24.8 \text{ to }40.7$ pp while reducing nominal-task success by at most $1.8$ pp.
\end{itemize}\vspace{-5pt}

\section{Related Work}
\subsection{Vision-Language-Action Models}
Vision-language-action (VLA) models couple pre-trained vision-language representations with visuomotor action generation \cite{kim2024openvla,intelligence2025pi_}. These models have demonstrated strong semantic generalization across manipulation and navigation tasks (e.g. aerial navigation \cite{wu2025vla}). Natural-language, however, primarily describes \emph{what} behavior is desired and need not precisely specify \emph{how} it should be executed (e.g. obstacle clearance margin). Recent work has therefore investigated safety-aware VLA training and evaluation \cite{li2026vision, zhang2026safevla,fan2026safevla}. Control barrier functions \cite{hu2025vlsa, english2026neuro} and MPC frameworks \cite{feng2025words} act as safety filters on VLA outputs. Temporal logic synthesis \cite{ravichandran2026safety} enforces safety during discrete symbolic planning. In contrast, we study formal-requirement-conditioned VLAs, where a temporal logic specification is provided as an additional policy input and may vary across deployments.\vspace{-5pt}

\subsection{VLA Post-training}
Supervised fine-tuning (SFT) adapts pre-trained robot policies by imitating demonstrated actions, but it does not explicitly distinguish between different-quality executions of the same task. Preference optimization \cite{meng2024simpo, liu2025survey} provides a mechanism for learning from such comparisons. Direct Preference Optimization (DPO) \cite{rafailov2023direct} optimizes a policy directly from preferred and dispreferred samples relative to a reference policy, while Identity Preference Optimization (IPO) \cite{azar2024general} provides an alternative pairwise objective designed to mitigate overfitting associated with preference-model assumptions.

Preference-based post-training has recently been extended to robot policies \cite{tian2024maximizing, chen2025fdpp}. GRAPE \cite{zhang2024grape} performs trajectory-wise preference optimization for VLA models using stage-wise costs generated from vision-language models and demonstrates alignment toward task completion, safety, and efficiency. FlowPRO \cite{wu2026flowpro} develops preference optimization specifically for flow-matching VLAs and introduces explicit regularization toward the reference policy. Related approaches \cite{zhang2026safevla} use constrained reinforcement learning to align VLAs with safety objectives. In these methods, the alignment objective is used to determine the behavior learned during post-training. Our setting differs in that the desired execution requirement is itself an input to the policy and may generalize at test time, similar to \cite{torne2026freeform}. We obtain supervision directly from formal temporal logic semantics. Post-training therefore does not bias the policy toward a single global notion of safety.\vspace{-5pt}

\subsection{Planning and Learning with Signal Temporal Logic}
Signal Temporal Logic (STL)~\cite{maler2004monitoring, donze2010robust} provides a language for specifying spatiotemporal properties of state trajectories and a robust semantics measuring the degree of satisfaction. STL has been incorporated into optimization-based \cite{raman2014model,pant2017smooth} and sampling-based \cite{marchesini2026sampling} planning and robot learning \cite{puranic2021learning}. Specification-conditioned reinforcement-learning methods~\cite{guo2024temporal} learn policies for temporal logic tasks. Generative approaches~\cite{meng2025telograf,meng2024diverse} condition diffusion or flow-matching models on temporal specifications. Recent work focuses on formal logic planning based on visual observations. S-MSP maps multi-view images and an STL specification to a trajectory and incorporates STL robust semantics into its training objective~\cite{ye2025bridging}, while Vision-TL-Action conditions a flow-matching trajectory generator on visual observations and a temporal logic syntax graph~\cite{liu2026vision}. These vision-based methods address specification-conditioned \emph{trajectory generation}, whereas our goal is to adapt a vision-language policy that repeatedly observes and acts in closed loop while jointly conditioning on a natural-language task and an STL requirement.

A complementary family of methods enforces formal requirements during deployment. STL robust semantics can guide generative-policy sampling through predicted future states~\cite{zoellner2026temporal}, and hierarchical frameworks can use STL for planning, monitoring, and replanning~\cite{torshizi2026step}. Such approaches perform additional optimization or reasoning during execution. Time-varying Control Barrier Functions \cite{lindemann2018control} act as STL-aware safety filters for planners, but their construction remains in state-space with difficult recursive feasibility guarantee. We use STL during post-training as both a policy condition and a source of trajectory-level supervision, so that requirement-dependent behavior is represented directly by the learned VLA policy. Our work connects specification-conditioned control with VLA post-training: a single visuomotor policy is adapted to preserve the NL task while responding to different temporal and geometric requirements supplied at test time.\vspace{-5pt}

\section{Preliminary}
Consider an off-the-shelf vision-language-action (VLA) model $\pi_\theta$, parameterized by $\theta$. Given a natural-language (NL) task $l$, at each decision time $t_d \in \mathbb{N}$, conditioned on a high-dimensional observation $o_{t_d}$, the policy samples a $K_a$-step action $\hat{a}^d \sim \pi_\theta(\cdot\mid o_{t_d}, l)$. At each control time $t \in \mathbb{N}$, an execution rule selects the applied action $\hat{a}_t$ from the VLA outputs available up to $t$. At each time $t$, a discrete-time system with state $s_t \in \mathbb{R}^n$ evolves under an unknown dynamics $f$ to reach the next-time state $s_{t + 1} \coloneq f(s_t, \hat{a}_t, w_t)$ where $w_t \in \mathcal{W}$ is some exogenous process disturbance from an unknown distribution. The VLA is invoked iteratively to produce a trajectory $s \coloneq (s_0, s_1, \hdots, s_H)$, where $H$ is the termination time. We say that $s$ is the trajectory rollout from $\pi_\theta$. 

Suppose the NL task $l$ specifies that a drone must circle around a table. A well-trained VLA may perform this nominal task, yet task completion alone does not guarantee compliance with additional formal requirements. For instance, the drone may be required to remain at least a prescribed distance from the table or return to base before a deadline. Such requirements can be difficult to express and evaluate precisely using NL alone. We therefore consider deployment-time formal requirements expressed using Signal Temporal Logic (STL). Existing VLA models generally do not account for such requirements during pre-training (i.e., large-scale training on diverse vision-language-action datasets for general perception, reasoning, and control capabilities). We study how to incorporate formal requirements through post-training (i.e., adapting an already pre-trained policy using additional requirement-aware objectives) while preserving the capabilities of the pre-trained policy.

\subsection{Signal Temporal Logic}
\label{subsec:stl}
Signal Temporal Logic (STL) is considered in \cite{maler2004monitoring, fainekos2009robustness}. Readers unfamiliar with temporal logic may simply regard an STL formula $\phi$ as a formal requirement on the system trajectory and skip the detailed syntax and semantics below.

Given a discrete-time trajectory $s$ and an STL formula $\phi$, we let $(s, \tau_0) \models \phi$ denote that $s$ satisfies $\phi$ at time $\tau_0$. Without loss of generality, we let $s \models \phi$ denote $(s, 0) \models \phi$ (i.e., $s$ satisfies $\phi$). The syntax of an STL formula follows 
\begin{align*}
    \phi \coloneq \text{True} \mid \pi^\mu \mid \neg \phi \mid \phi_1 \wedge \phi_2 \mid \phi_1 U_{[a, b]} \phi_2
\end{align*}
where $\pi^\mu: \mathbb{R}^n \rightarrow \{\text{True}, \text{False}\}$ is a predicate such that $\pi^\mu(s_t) \coloneq \text{True}$ if $\mu(s_t) \ge 0$ and $\pi^\mu(s_t) \coloneq \text{False}$ otherwise, where $\mu:\mathbb{R}^n \rightarrow \mathbb{R}$ is a user-specified function. Predicates are atomic elements of an STL and can be used to express spatial relations (e.g. $\mu(s_t) \coloneq \|s_t - O\|_2 - d_{\min}$ expresses obstacle avoidance, where $O$ is some obstacle location and $d_{\min}$ is some clearance margin). The symbols $\neg$ and $\wedge$ are Boolean operators denoting \textit{negation} (not) and \textit{conjunction} (and) respectively. The symbol $U_{[a, b]}$ is the temporal operator \textit{until} with $a \le b$ and $a, b \in \mathbb{N}$, which encodes that $\phi_1$ has to be true until $\phi_2$ is true at some future time in $[a, b] \cap \mathbb{N}$. From these operators, we can derive the operators for \textit{disjunction} by $\phi_1 \vee \phi_2 \coloneq \neg(\neg\phi_1 \wedge \neg \phi_2)$, for \textit{implication} by $\phi_1 \implies \phi_2 \coloneq \neg \phi_1 \vee \phi_2$, for \textit{eventually} by $F_{[a, b]}\phi \coloneq \text{True}\;U_{[a, b]}\;\phi$, and for \textit{globally} by $G_{[a, b]}\phi \coloneq \neg F_{[a, b]}\neg \phi$. \fullswitch{While the semantics of STL are standard \cite{maler2004monitoring}, we recall them in Appendix~\ref{app:semantics_stl}. We also define in Appendix~\ref{app:semantics_stl} the robust semantics $\rho^\phi(s, \tau_0)$ of STL \cite{donze2010robust, fainekos2009robustness}, which evaluate how robustly $\phi$ is satisfied vs. violated.}{The semantics of STL are standard \cite{maler2004monitoring}. The robust semantics $\rho^\phi(s, \tau_0)$ of STL are defined in \cite{donze2010robust, fainekos2009robustness}, which evaluate how robustly $\phi$ is satisfied vs. violated.} Importantly, $(s, \tau_0) \models \phi$ if $\rho^\phi(s, \tau_0) > 0$ and a larger $\rho^\phi(s, \tau_0)$ denotes that the trajectory is more satisfied \cite{fainekos2009robustness}. We use $\rho^\phi(s)$ to denote $\rho^\phi(s, 0)$. Given $s$ and $\tau_0$, one needs a sufficient termination $H \coloneq H^\phi$ to evaluate satisfaction\fullswitch{. We recall $H^\phi$ in Appendix~\ref{app:semantics_stl}.}{ \cite{sadraddini2015robust}.}\vspace{-5pt}

\subsection{Problem Formulation}
We study how to post-train a pre-trained VLA for deployment-time STL requirements while preserving its NL task-following capability. Given a pre-trained VLA, which we call the base policy, we target its satisfaction of an STL requirement at test time. We formalize Problem \ref{prob:overall}.

\begin{problem}
    \label{prob:overall}
    Consider a well-trained\footnote{The model generates rollouts that reasonably satisfy the NL tasks.} VLA model $\pi_\theta$ with a fixed NL specification distribution $\mathcal{L}$ and environment distribution $\mathcal{E}$ in which the dynamics $f$ operates. Design a post-train procedure $ \tilde{\theta} = \mathcal{A}(\theta)$ such that, at inference time, $\pi_{\tilde{\theta}}(\cdot\mid o, l, \phi)$ jointly conditions on the observation $o$ from environment $\mathcal{E}$, NL task $l \sim \mathcal{L}$, and STL requirement $\phi$, and produces a trajectory $s$ satisfying both $s \models \phi$ and the nominal task $l$.
\end{problem}

Problem \ref{prob:overall} poses several challenges: 1) Post-training must avoid significantly altering the base policy’s rollout distribution, which could degrade NL task performance. 2) Learning requirement-dependent behavior from structured STL formulas requires capturing their spatiotemporal semantics, which may be difficult to acquire reliably through SFT alone. 3) STL semantics are global functionals over long-horizon trajectories, and the optimization must avoid reducing a trajectory-level requirement to independent action-chunk preferences. To provide sufficient demonstrations for our post-training procedure $\mathcal{A}$, we consider the following data assumption.
\begin{assumption}
\label{ass:data}
    We assume access to an in-domain dataset $D \coloneq \{p^{(1)}, \hdots, p^{(N_D)}\}$, where $p^{(i)} \coloneq (s^{(i)}, a^{(i)}, l^{(i)}, o^{(i)})$ with $s^{(i)}$ denoting a trajectory sample, $a^{(i)}$ denoting the action sequence, $l^{(i)}$ denoting the corresponding NL specification, and $o^{(i)}$ denoting the observation sequence leading to $s^{(i)}$.
\end{assumption}
The purpose of Assumption \ref{ass:data} is to provide in-distribution datasets, from which an offline monitor enables efficient STL learning during the post-training stage (as we show in Section \ref{sec:method}). Trivially, one can directly sample $D$ from the dataset used in training the base model or directly rollout $\pi^\theta$ to collect data that are roughly in-distribution. The STL specification enforced at test-time must not conflict with the NL specification, in which case Problem \ref{prob:overall} is infeasible.

\begin{assumption}
    In Problem \ref{prob:overall}, the NL specification $l$ is not in conflict with the STL requirement $\phi$. 
\end{assumption}

To address Problem \ref{prob:overall}, we propose Logic-VLA, the requirement-conditioned policy resulting from our post-training procedure $\mathcal{A}$. The model augments a pre-trained VLA with an STL conditioning pathway, while $\mathcal{A}$ determines how its parameters are adapted from $\theta$ to $\tilde{\theta}$.\vspace{-10pt}

\section{Logic-VLA}
\label{sec:method}
\subsection{Logic-Conditioned Supervision Construction}
\label{subsec:data_curation}
Given the in-distribution dataset $D = \{p^{(i)}\}_{i = 1}^{|D|}$ from Assumption~\ref{ass:data}, we construct the STL-conditioned supervision used for post-training. We generate a bank of candidate STL specifications and construct satisfying demonstrations and matched satisfying–violating trajectory pairs which provide supervision for the two post-training stages in Section \ref{sec:post-training}.

\paragraph{Candidate formula bank} Let $\mathcal{P}_{\text{STL}}$ denote a distribution over well-formed STL formulas under the grammar introduced in Section \ref{subsec:stl}. A draw from $\mathcal{P}_{\text{STL}}$ specifies both the syntactic structure of the formula and its predicate and temporal parameters. We sample a finite candidate formula bank $\Phi \coloneq \{\phi^{(1)}, \hdots, \phi^{(N_\phi)}\}$ where $\phi^{(j)} \sim \mathcal{P}_{\text{STL}}$. At inference, we likewise consider an STL requirement $\phi \sim \mathcal{P}_{\text{STL}}$ which need not belong to $\Phi$. The goal is therefore to learn a specification-conditioned policy that generalizes from the formulas observed during post-training to unseen formulas drawn from the same specification distribution. \fullswitch{The concrete formula templates and parameter distributions used to instantiate $\mathcal{P}_{\text{STL}}$ in our evaluation of Section \ref{sec:eval} are provided in Appendix \ref{sec:stl_generator}.}{}

\paragraph{Satisfying demonstrations and preference pairs} Our two-stage post-training procedure requires two complementary forms of supervision. Stage 1 learns specification-conditioned behavior by imitating executions that satisfy the given STL requirement, while Stage 2 requires paired satisfying and violating executions so that the policy can learn which behavior should be preferred under the same task and specification. Accordingly, given the rollout dataset $D$ and candidate formula bank $\Phi$, we use a monitor \cite{nivckovic2020rtamt} to evaluate each specification on the collected trajectories. Robust semantics are used to identify executions that clearly satisfy or violate a specification, and rollouts are matched only when they share the same nominal NL task and sufficiently comparable environment and initial conditions. From the satisfying rollouts, we extract STL-conditioned action demonstrations $D^+ = \{(\xi_{i,j, \tau}, a_\tau^{(i)})\}$ where $\xi_{i,j,\tau}=(o^{(i)}_\tau,l^{(i)},\phi^{(j)})$ is the STL-conditioned context where $\phi^{(j)} \in \Phi$ is the STL specification satisfied by rollout $i$ and $a_\tau^{(i)}$ is the demonstrated action chunk extracted at action-window index $\tau$. For stage 2, we construct the trajectory-level preference dataset $P = \{(p_i^+, p_i^-, l_i, \phi_i)\}$ where $p_i^+$ and $p_i^-$ are matched rollouts of the same nominal task $l_i$ and are under comparable environment and initial conditions, with $p_i^+$ clearly satisfying and $p_i^-$ clearly violating the same STL $\phi_i$. \fullswitch{The precise rollout-grouping criteria and pair-selection procedure are provided in Appendix \ref{sec:stl_generator}.}{}\vspace{-5pt}
\begin{figure*}
    \centering    
    \includegraphics[width=0.9\textwidth]{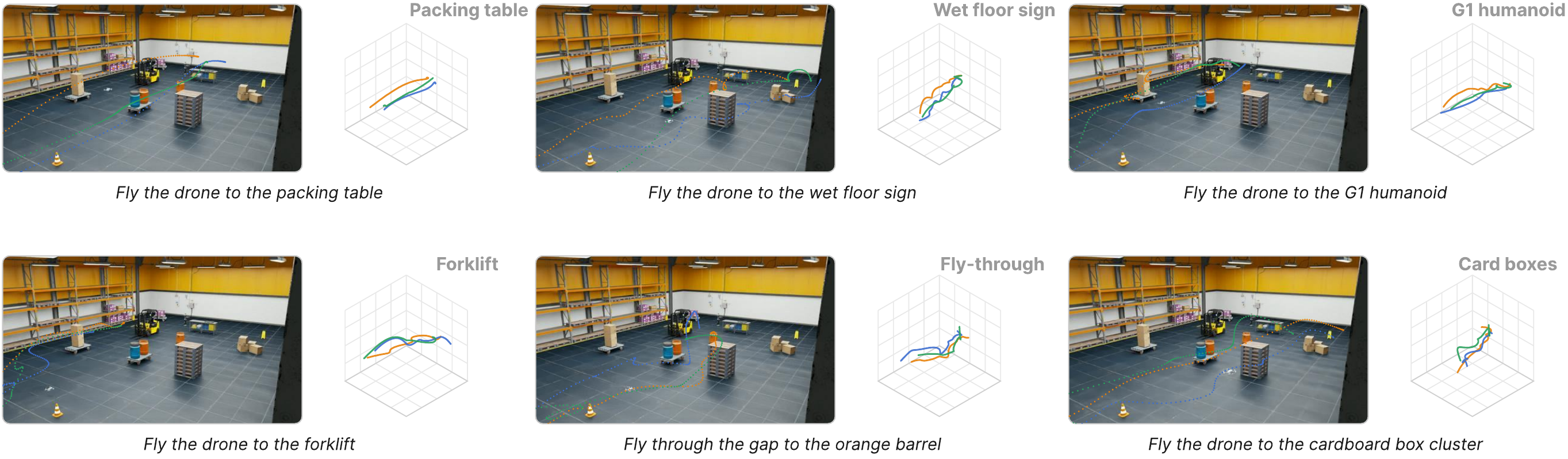}
    \caption{\textbf{Visualization of the six natural-language navigation tasks.}
For each task, we show several example demonstration trajectories from the dataset $D$ in one of
the ten randomized warehouse environments used in our experiments.}
    \label{fig:task}
    \vspace{-15pt}
\end{figure*}
\subsection{Logic Encoder}
\label{subsec:stl_encoder}
One key design component is a logic encoder that maps an STL specification into the VLA token space for integration with multimodal sensory representations. Specifically, it encodes each formula as a graph-structured representation and is pre-trained using STL robust semantics to capture the predicate, temporal, and logical structure of the STL formulas.

\paragraph{STL encoder} We adopt the syntax-graph backbone of TeLoGraF~\cite{meng2025telograf}. An STL formula  $\phi$ is represented as a directed graph
$G_\phi=(V_\phi,E_\phi)$, whose nodes encode Boolean, temporal, and comparison operators, with edges directed from operands to their parent operators. Each node $v \in V_\phi$ is associated with a feature vector $\mathbf{x}_v = [\kappa_v, t_v^s, t_v^e, \mathbf{s}_v, c_v, \alpha_v]$, where $\kappa_v \in \{\textsc{And},\textsc{Or},\textsc{Not},
\mathbf{F},\mathbf{G},\leq,\geq\}$ identifies the operator type. For a bounded temporal operator over the time interval $[a, b]$ where $a, b$ are temporal bounds expressed in consistent units, we set $(t_v^s, t_v^e) = (a, b)$ and use a sentinel value $(-1, -1)$ otherwise. Recall that each atomic predicate $\pi^\mu$ is defined by a predicate function $\mu:\mathbb{R}^n\rightarrow\mathbb{R}$, with $\pi^\mu(s_t)=\mathrm{True}$ iff $\mu(s_t)\geq0$. We consider a finite vocabulary of scalar signals $\mathcal{G}\coloneq \{g_1,\ldots,g_M\}$ and parameterize each predicate by a signal $g_q\in\mathcal{G}$, a threshold $c\in\mathbb{R}$, and a comparison relation $\bowtie\in\{\leq,\geq\}$. In particular, the predicate $g_q(s_t)\bowtie c$ is represented by $\mu(s_t)\geq0$, with $\mu(s_t)\coloneq g_q(s_t)-c$ for $\bowtie=\geq$ and $\mu(s_t)\coloneq c-g_q(s_t)$ for $\bowtie=\leq$. We encode the identity of $g_q$ through $\mathbf{s}_v\in \{-1,1\}^M$, whose $q$-th entry is $1$ and all remaining entries are $-1$; for non-predicate nodes, we set $\mathbf{s}_v=-\mathbf{1}$. The scalar $c_v=c$ records the predicate threshold, with $c_v \coloneq 0$ for non-predicate nodes. We retain $\alpha_v$ as an auxiliary bookkeeping field and set $\alpha_v = -1$ throughout. For instance, consider a two-dimensional state $s_t = (x_t, y_t) \in \mathbb{R}^2$ and signal vocabulary $\mathcal{G} = \{g_x, g_y\}$, where $g_x(s_t) \coloneq x_t$ and $g_y(s_t) \coloneq y_t$. Their signal encodings are $(1, -1)$ and $(-1, 1)$, respectively. The predicate $x_t \ge 2$ is schematically represented by a predicate node with $\mathbf{x}_v = [\ge, -1, -1, (1, -1), 2, -1]$. Unlike TeLoGraF \cite{meng2025telograf}, which encodes object geometry in the STL graph and conditions trajectory generation on the initial state, our encoder represents only the symbolic requirement. Variables needed to evaluate a predicate, such as obstacle pose or ego–obstacle distance, may be included in the offline state trajectory $s$ used to compute STL semantics, but are not explicit inputs to the VLA during policy post-training or deployment. Instead, the specification is grounded through the VLA’s observations, decoupling the logic representation from an object configuration. We adopt TeLoGraF \cite{meng2025telograf} for the syntax-graph encoder architecture, using child-to-parent GCN message passing followed by mean pooling for a formula-level representation $E_{\text{STL}}(\phi) \coloneq 
\text{MeanPool}(\text{GCN}(G_\phi, X_\phi))$, where $X_\phi$ stacks the node features $\{\mathbf{x}_v\}_{v \in V_\phi}$. The encoder representation $E_{\text{STL}}(\phi)$ is independent of the observation.

\paragraph{STL encoder pre-training} While the graph encoder preserves the syntactic structure of an STL formula, we further pre-train $E_{\text{STL}}$ to capture the STL semantics. We construct a set of trajectory-formula pairs $S_{\text{pre}} \subseteq D \times \Phi$\fullswitch{ where the precise sampling and selection procedure is provided in Appendix \ref{sec:stl_generator}}. For each pair $(p^{(i)}, \phi^{(j)}) \in S_{\text{pre}}$, we compute the robust semantics $\rho^{(i, j)} \coloneq \rho^{\phi^{(j)}}(s^{(i)})$. We use these pairs to train $E_{\text{STL}}$ through an auxiliary robust semantics prediction objective. An auxiliary trajectory encoder $E_{\text{traj}}$ maps $s^{(i)}$ to a latent representation, which is combined with $E_{\text{STL}}(\phi^{(j)})$ by an MLP regression head $g_\psi$. We optimize $\mathcal{L}_{\mathrm{pre}}
    =
    \mathbb{E}_{(p^{(i)},\phi^{(j)}) \in S_{\mathrm{pre}}}
    \left[
        H_{\delta}\!\left(
            \hat{r}^{(i,j)}
            -
            \tanh\!\left(\rho^{(i,j)}/c\right)
        \right)
    \right],$ where $H_{\delta}$ is the Huber loss with threshold $\delta$, and
$c>0$ is a scale parameter used to normalize the robust semantics. Predicting the robust semantics encourages the embedding to encode the predicate thresholds, temporal bounds, and logical composition that determine how a trajectory satisfies or violates a formula. After pre-training, we discard $E_{\text{traj}}$ and $g_\psi$. The pretrained $E_{\text{STL}}$ is then jointly fine-tuned with the policy during post-training. For downstream VLA conditioning, we project the pre-trained formula representation into a sequence of $N_{\text{spec}}$ specification tokens, $Z_\phi \coloneq
    \operatorname{reshape}
    \!\left(
        W_pE_{\text{STL}}(\phi)+b_p
    \right)
    \in\mathbb{R}^{N_{\text{spec}} \times d_c},$ where $d_c$ is the hidden dimension of the VLA conditioning stream, and $W_p$ and $b_p$ are learned parameters. We append $Z_\phi$ to the VLM prefix after the image and NL tokens, where bidirectional attention lets perception condition on the specification.\vspace{-7pt}
\subsection{Architecture and Two-Stage Post-training}
\label{sec:post-training}
We choose $\pi_{0.5}$ \cite{intelligence2025pi_} as the backbone of Logic-VLA but remark that the post-train recipe can be used for other flow-matching policies. The latent representation $Z_\phi$ is appended into the vision-language prefix, while the underlying backbone architecture is unchanged and initialized with the pre-trained weights. The full architecture is shown in Figure \ref{fig:architecture}. Starting from the semantically pre-trained STL encoder, we adapt the policy in two stages using the datasets curated in Section \ref{subsec:data_curation}. \emph{Stage 1} is an STL-conditioned SFT on the satisfying demonstrations $D^+$, establishing specification-conditioned behavior through imitation. \emph{Stage 2} performs trajectory-level preference optimization on the satisfying-violating pairs $P$, teaching the policy to prefer executions that satisfy the given STL requirement over comparable executions that violate it.

\paragraph{Stage 1: STL-conditioned supervised fine-tuning} For a demonstration $(\xi, a) \in D^+$, where $\xi = (o, l, \phi)$ denotes the STL-conditioned context, and $a$ is the demonstrated action chunk, we optimize the standard conditional flow-matching objective of the $\pi_{0.5}$ action expert \cite{intelligence2025pi_}. Since $D^+$ contains action demonstrations from trajectories satisfying their associated STL requirements, this stage adapts the policy to imitate satisfying behavior while conditioning on the STL formula. The STL encoder and VLA parameters are jointly fine-tuned, with no additional STL-specific loss. Let $\pi_{\theta_\text{SFT}}$ denote the resulting policy. We initialize Stage 2 as $\tilde{\theta} \leftarrow \theta_\text{SFT}$ and retain a frozen copy as the reference policy $\pi_{\text{ref}} = \pi_{\theta_\text{SFT}}$.
\begin{table*}[t]
    \centering
    \scriptsize
    \setlength{\tabcolsep}{3.0pt}
    \renewcommand{\arraystretch}{1.08}

    \resizebox{\textwidth}{!}{%
    \begin{tabular}{lcccc|cccc|cccc}
        \hline
        &
        \multicolumn{4}{c|}{Seen}
        &
        \multicolumn{4}{c|}{Unseen Parameter}
        &
        \multicolumn{4}{c}{Unseen Structure} \\
        \cline{2-13}

        Method
        & STL Sat. $\uparrow$
        & $\rho$ Mean $\uparrow$
        & $\rho_{\min}$ $\uparrow$
        & NL Task $\uparrow$

        & STL Sat. $\uparrow$
        & $\rho$ Mean $\uparrow$
        & $\rho_{\min}$ $\uparrow$
        & NL Task $\uparrow$

        & STL Sat. $\uparrow$
        & $\rho$ Mean $\uparrow$
        & $\rho_{\min}$ $\uparrow$
        & NL Task $\uparrow$ \\
        \hline

        Base
        & 41.3
        & $-0.35{\pm}1.33$
        & -9.11
        & \textbf{90.5}

        & 50.0
        & $-0.03{\pm}1.26$
        & -5.80
        & \textbf{93.9}

        & 56.8
        & $0.07{\pm}1.51$
        & -6.98
        & \textbf{89.3}
        \\

        STL-SFT
        & 61.7
        & $0.49{\pm}1.57$
        & -5.09
        & 87.2

        & 59.5
        & $0.34{\pm}1.52$
        & -5.05
        & 91.2

        & 64.5
        & $0.71{\pm}1.86$
        & -4.64
        & 85.2
        \\

        Smooth Robust Semantics ($1\times$)
        & 76.2
        & $1.62{\pm}2.92$
        & -5.72
        & 68.5

        & 71.1
        & $1.24{\pm}2.51$
        & -4.94
        & 75.9

        & 77.5
        & $2.18{\pm}3.26$
        & -4.92
        & 68.6
        \\

        Smooth Robust Semantics ($2\times$)
        & 78.8
        & \textbf{$2.33{\pm}3.80$}
        & -5.79
        & 45.0

        & 72.1
        & \textbf{$1.81{\pm}3.19$}
        & -7.31
        & 47.4

        & 81.8
        & \textbf{$3.34{\pm}4.24$}
        & -3.82
        & 42.0
        \\

        \rowcolor{blue!8}
        \textbf{Logic-VLA}
        & \textbf{82.0}
        & $0.81{\pm}1.33$
        & \textbf{-1.62}
        & 89.0

        & \textbf{74.8}
        & $0.89{\pm}1.64$
        & \textbf{-3.79}
        & 92.2

        & \textbf{82.0}
        & $1.24{\pm}1.71$
        & \textbf{-2.46}
        & 87.5
        \\

        \hline
    \end{tabular}%
    }

    \caption{\textbf{Evaluation Results.} Comparison of Results across the baselines in \emph{Seen}, \emph{Unseen Parameter}, and \emph{Unseen Structure}.}
    \vspace{-15pt}
    \label{tab:main_results}
\end{table*}
\paragraph{Stage 2: trajectory-level preference optimization} Stage 2 trains on the matched satisfying-violating rollout pairs $P$. For pair $i$, let $p_i^+$ and $p_i^-$ denote the satisfying and violating rollouts associated with the same NL task $l_i$ and STL requirement $\phi_i$. At temporal window $\tau$, let $a_{i, \tau}^+$ and $a_{i, \tau}^-$ denote the action chunks extracted from $p_i^+$ and $p_i^-$ respectively. We evaluate both candidates under the shared conditioning context $\xi_{i, \tau} \coloneq (o_{i, \tau}^+, l_i, \phi_i)$, where $o^+_{i, \tau}$ is the observation from the satisfying rollout. 

We adopt Identity Preference Optimization (IPO) \cite{azar2024general}, which learns a finite-margin preference between preferred and rejected samples relative to a reference policy. For likelihood-based policies, this margin is expressed via reference-relative log-policy ratios. However, the $\pi_{0.5}$ action expert is parameterized by conditional flow matching, for which exact action likelihood ratios are computationally expensive to evaluate. Following the likelihood-ratio surrogate of \cite{mcallister2026flow}, we therefore use differences in conditional flow-matching losses as a surrogate for the reference-relative log-policy ratios. Because STL satisfaction is a trajectory-level property rather than an action-chunk-level property, we aggregate the policy score across a collection $W_i$ of temporal windows sampled from pair $i$. Let $\ell_{\text{FM}}(\nu; a_\tau, \xi_\tau, \gamma, \epsilon)$ denote the standard conditional flow-matching loss of policy parameters $\nu$ for action chunk $a_\tau$ under context $\xi_\tau$, flow time $\gamma$, and noise $\epsilon$. For $\nu \in \{\tilde{\theta}, \theta_\text{SFT}\}$, define the trajectory-level flow-matching loss $\bar{\ell}_{\nu,i}^{\pm}\coloneq\frac{1}{|W_i|}\sum_{\tau\in W_i}\ell_{\text{FM}}\left(\nu; a_{i,\tau}^{\pm}, \xi_{i,\tau}, \gamma_i,\epsilon_i\right)$. The same $(\gamma_i,\epsilon_i)$ realization is used across the temporal
windows and both current/reference evaluations of pair $i$\fullswitch{, while the window-sampling procedure is described in Appendix \ref{app:arch}}{}. Following the flow-matching likelihood surrogate, we define the reference-relative trajectory scores $ q_i^{\pm}= \bar{\ell}_{\theta_{\text{SFT}},i}^{\pm}-\bar{\ell}_{\tilde{\theta},i}^{\pm}$. The corresponding  preference margin is $\Delta_i=\beta\left(q_i^{+}-q_i^{-}\right)$, where $\beta > 0$ controls the scale of the preference margin. Following IPO, we optimize this margin toward a finite target $c$ using a Huberized IPO-style finite-margin objective, $\ell_{\mathrm{IPO},i}= H_{\delta'}(\Delta_i-c)$, where $H_{\delta'}(\cdot)$ denotes the Huber loss with width
$\delta' > 0$. \fullswitch{The target $c$ controls the desired finite preference margin with calibration details in Appendix \ref{app:arch}.}

A relative preference objective can increase the satisfying–violating margin even if the current policy becomes worse at fitting the satisfying rollout, provided that the violating rollout degrades more. To prevent this, we introduce a one-sided preferred-rollout anchor $A_i=
    \left[
        \bar{\ell}_{\tilde{\theta},i}^{+}
        -
        \mathrm{sg}
        \left(
            \bar{\ell}_{\theta_\mathrm{ref},i}^{+}
        \right)
    \right]_+ $, where $sg(\cdot)$ denotes stop-gradient. The anchor is active only when the current policy incurs a larger flow-matching loss on the satisfying rollout than the frozen Stage-1 reference policy. The complete Stage 2 objective is $\mathcal{L}_{\mathrm{pref}}\coloneq \mathbb{E}_{i}
    \left[
        \ell_{\mathrm{IPO},i}
        +
        \lambda A_i
    \right]$, where $\lambda \ge 0$ controls the strength of the preferred-rollout anchor. The preference term learns the STL-dependent ordering between satisfying and violating executions, while the anchor preserves the satisfying behavior acquired during Stage 1.
\section{Evaluation}
\label{sec:eval}
We organize our empirical evaluation around three research questions.
\textbf{1) Q1: Requirement satisfaction and task preservation.}
Can Logic-VLA improve satisfaction of STL requirements while
preserving the nominal NL task compared to pure STL-conditioned imitation or robust semantics driven optimization?
\textbf{2) Q2: Specification generalization.}
Does the learned policy generalize to STL formulas unseen during post-training?
\textbf{3) Q3: Ablative design choices.}
What is the effect of pre-training the STL encoder and choosing syntax-graph STL encoding vs. prompting via the NL channel? We first describe our simulation design, metrics, and baselines. We address Q1 and Q2 in Section \ref{subsec:main_results} and Q3 in Section \ref{subsec:ablations}.
\subsection{Experimental Setup}
\paragraph{Environment and tasks}
We instantiate and evaluate Logic-VLA in a closed-loop quadcopter
navigation setting using NVIDIA Isaac Sim~5.1. We consider ten
randomized photorealistic warehouse environments and six
natural-language navigation tasks (shown in Figure \ref{fig:task}) centered on reaching ordered target
regions. \fullswitch{To collect training data $D$ outlined in Assumption \ref{ass:data}, we propose an efficient automatic certifiable trajectory generation pipeline, CRATE, which we discuss in detail in Appendix \ref{app:crate}. CRATE generates $|D| = 3000$ collision-free, dynamically feasible
reference trajectories spanning these environment--task combinations.}{We collect $|D| = 3000$ collision-free dynamically feasible reference trajectories spanning these environment--task combinations.}
We execute these trajectories with a DJI Mavic~2 Pro model in simulation
and record the corresponding actions (absolute drone position) together with synchronized visual observations from both an egocentric onboard camera and a third-person room-view camera, which are provided to the policy as visual inputs in the navigation demonstrations. Our control procedure is via position update through Isaac Sim API calls, tracking the predicted positions.

Starting from a pre-trained
$\pi_{0.5}$ base checkpoint, we fine-tune the policy on the collected demonstrations $D$ without STL conditioning. The resulting STL-blind navigation policy serves as
the common initialization for all post-training
baselines. We adapt the $2B$ vision--language backbone and $300M$ action expert using LoRA with ranks $16$ and $32$ ($\alpha=16$ and $32$), respectively, while fully fine-tuning the vision encoder and action input/output projections throughout the baselines and training procedures. During evaluation, the policy executes $40$ actions before
re-querying the policy from the latest observation in closed loop at a control frequency of $10$ Hz. \fullswitch{The construction details of the environments, tasks, and trajectories are in Appendix~\ref{app:arch}.}{}
\begin{figure*}[t]
    \centering
    \includegraphics[width=0.8\textwidth]{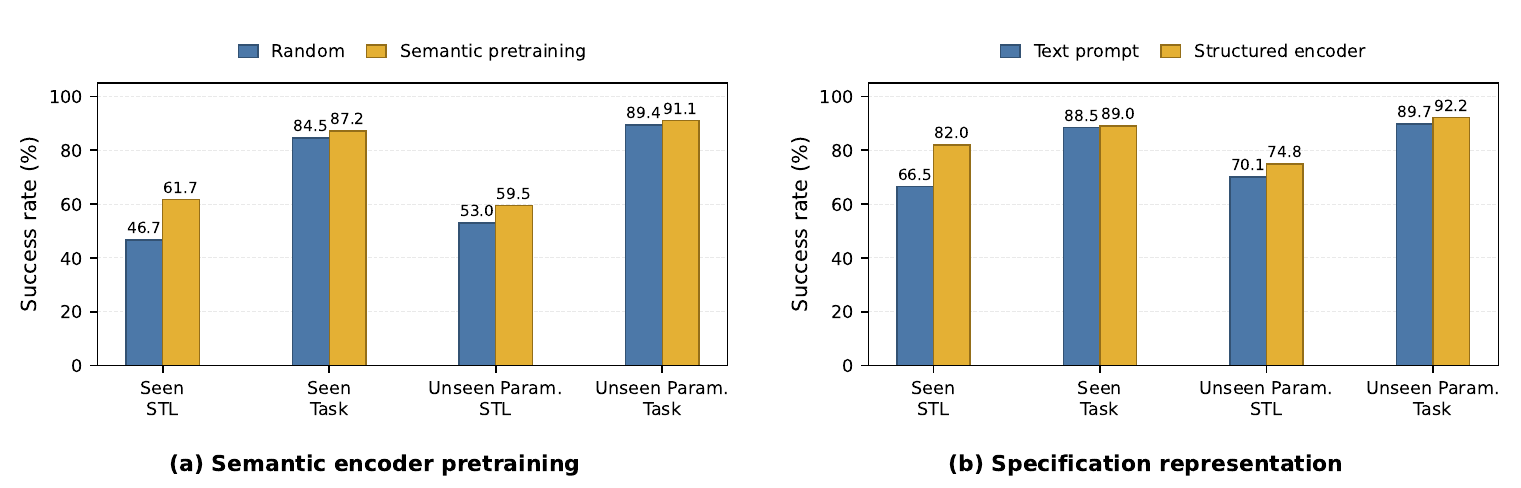}
    \caption{\textbf{Ablation Results.} (a) Semantic pre-training vs. random encoder initialization. (b) Syntax-graph encoding vs. textual prompting. Results are shown on \textit{Seen} and \textit{Unseen Parameter} settings for STL satisfaction and NL task success.}
    \vspace{-15pt}
    \label{fig:stl_encoder_ablations}
\end{figure*}

\paragraph{Evaluation procedure}
Consider a fixed initial state distribution $\mathcal{X}$. We start with a templated STL generation procedure using a predefined distribution $\mathcal{P}_{\text{STL}}$, from which we sample the formula bank $\Phi$. We instantiate $\mathcal{P}_{\mathrm{STL}}$ from 90 \textit{structural templates} obtained by enumerating temporal compositions
$\{F,G,FG,GF,FGF,GFG\}$ and Boolean combinations using $\land$ and $\lor$. For each structure, STL formulas are generated by sampling its
predicate (within the family of $x, y, z \bowtie c$) and temporal parameters from predefined ranges, where $s_t \coloneq (x_t, y_t, z_t)$ denotes the drone position. We define a \emph{formula-group} as a set of trajectory–formula pairs sharing the same NL task, environment, and STL formula, with trajectory initial states lying in a local neighborhood within the support of $\mathcal{X}$. We use these groups to construct the supervision described
in Section~\ref{subsec:data_curation}: satisfying trajectories contribute
to $D^+$, while matched satisfying-violating trajectory pairs contribute
to $\mathcal{P}$. This grouping controls task- and environment-induced
variation while preserving execution-level diversity. \fullswitch{See Appendix \ref{sec:stl_generator} for the STL generation and grouping details.}{}

After filtering and grouping, we obtain $1449$ formula-groups generated from $1224$ distinct formulas spanning $87$ STL structures. These groups provide $8886$ satisfying rollouts, from which we construct the dataset $\mathcal{D}^{+}$. Among them, $6754$ satisfying instances admit at least one matched
violating counterexample, producing $13494$ preference-pair
instances in $\mathcal{P}$ for Stage~2.\fullswitch{ The details of the selection procedure are in Section \ref{sec:stl_generator}.}{} Using datasets $D^+$ and $P$, we apply the procedure in Section \ref{subsec:stl_encoder} and \ref{sec:post-training} to obtain policy $\pi_{\tilde{\theta}}$. We evaluate $\pi_{\tilde{\theta}}$ and baselines under three test settings. Each \emph{evaluation entry} specifies an NL task $l$, an STL formula $\phi$,
and a task environment. For each entry, we sample $10$ initial states
$s_0 \sim \mathcal{X}$ from the corresponding initial-state distribution
and execute one closed-loop rollout from each state. \emph{In the first test setting, we evaluate over the STL formulas seen during post-training}: \emph{Seen} contains $60$ evaluation entries sampled from the formula-groups used for post-training. \emph{In the second setting, we test the ability of our policy in generalizing to unseen parameters under STL structures used for post-training and STL encoder pre-training:} \emph{Unseen Parameter} contains $99$ evaluation entries, covering all $87$ STL structures observed during post-training but with newly sampled predicate and temporal parameters held out from training. \emph{In the last setting, we evaluate over STL structures unseen during post-training:} \textit{Unseen Structure} contains 56 evaluation entries covering 49 STL formulas
instantiated from 3 structural templates held out from both post-training and
STL encoder pre-training.

\paragraph{Metrics}
For each evaluation entry, we execute 10 closed-loop rollouts.
Our primary requirement metric is the \emph{STL satisfaction rate},
defined as the fraction of rollouts satisfying the provided
specification\footnote{We adopt the convention that zero robust semantics constitutes satisfaction; this differs from the strict positive robust semantics only when $\rho^\phi(s) = 0$.}, i.e., $\rho^\phi(s) \ge 0$.
We additionally report the mean STL robust semantics and its standard deviation,
as well as the minimum robust semantics $\rho_{\min}$ across all evaluated
rollouts, to characterize both average and worst-case requirement
satisfaction. To measure preservation of the nominal natural-language task, we report
\emph{NL task success}, where a rollout is successful if it reaches all target regions in the prescribed order. \fullswitch{Auxiliary metrics such as the drone collision rate is reported in Appendix \ref{app:arch}.}{}
\paragraph{Compared methods}
All STL-conditioned baselines are initialized
from the same STL-blind policy. Their STL graph encoders are initialized from the same
pre-trained checkpoint; because the VLM conditioning stream
has a different hidden width from encoder pre-training, the token
projection is freshly initialized and trained during policy
post-training.

\begin{itemize}
\item \emph{Base} is the task-adapted STL-blind $\pi_{0.5}$ policy and serve as the initialization for all STL-aware methods.

\item \emph{STL-SFT} fine-tunes the base policy on the satisfying
demonstrations $D^{+}$ using only the standard conditional
flow-matching imitation objective described in
Sec.~\ref{sec:post-training}. It tests whether
specification-conditioned imitation alone is sufficient to induce
requirement-aware behavior.

\item \emph{Smooth Robust Semantics} uses the same initialization and $D^+$ as STL-SFT, but augments
the flow-matching objective with direct maximization of differentiable
STL robust semantics, where $\mathcal{L}_{\mathrm{smooth}} =\mathcal{L}_{\mathrm{FM}} + \lambda_{\rho}\mathcal{L}_{\rho}$ with $\mathcal{L}_{\rho} =-\mathbb{E}\left[\widetilde{\rho}(\hat{s},\phi)\right]$. Here,  $\widetilde{\rho}$ denotes the differentiable smooth robust
semantics \cite{pant2017smooth} evaluated on the reconstructed trajectory $\hat{s}$. We evaluate a nominal robust semantics loss weight $\lambda_\rho$ ($1\times$) and a doubled
weight ($2\times$); the two variants are otherwise identical.\fullswitch{ The
smooth-semantics construction and optimization details are provided in Appendix \ref{app:arch}.}

\item \emph{Logic-VLA} uses STL-SFT as its first stage and subsequently performs
trajectory-level preference optimization on the matched
satisfying-violating pairs $\mathcal{P}$. Stage~2 optimizes the
reference-relative IPO margin together with the preferred-trajectory
anchor described in Sec.~\ref{sec:post-training}.
\end{itemize}\vspace{-5pt}

\subsection{Requirement Satisfaction and Generalization}
\label{subsec:main_results}
The results are shown in Table \ref{tab:main_results}. Noticeably, conditioning the policy on STL through SFT already provides a substantial improvement over the STL-blind base policy for all three settings. Direct smooth robust semantics optimization further increases requirement satisfaction, but produces a pronounced trade-off with the nominal NL task. Raising $\lambda_\rho$ from $1\times$ to $2\times$ increases seen-specification satisfaction from $76.2\%$ to $78.8\%$, while task success decreases from $68.5\%$ to $45.0\%$. The same
pattern appears on unseen formulas. This tradeoff is expected: Compared to preference optimization, directly maximizing robust semantics alters the policy behavior beyond what is necessary to satisfy the STL requirements. In all three settings, Logic-VLA achieves the highest observed STL satisfaction while retaining nominal-task success at a level comparable to the base and STL-SFT policies, demonstrating that satisfying-violating sample pairs provide supervision about which executions are favored with information unavailable to positive-only imitation. Across all three settings, Logic-VLA improves satisfaction over the STL-blind base policy by
$24.8$ to $40.7$ percentage points (pp), while the corresponding reduction in
task success is at most $1.8$ pp. The generalization results show that Logic-VLA responds to the formal structure and parameters of the requirement rather than only memorizing the
specifications observed during post-training.\vspace{-5pt}

\subsection{Ablation Studies}
\label{subsec:ablations}
To address \textbf{Q3}, we ablate two designs:
semantic initialization of the STL encoder and structured syntax-graph
encoding versus direct textual prompting. Our ablations are under the same evaluation setup as Section \ref{subsec:main_results} with the \textit{Seen} and \textit{Unseen Parameter} test settings. Results are in Figure \ref{fig:stl_encoder_ablations}.\fullswitch{ Additional ablations are provided in Appendix \ref{app:arch}.}

\paragraph{Effect of STL encoder pre-training}
We evaluate whether pre-training the STL encoder via robust semantics in Section \ref{subsec:stl_encoder} provides a useful
initialization for policy learning. Under otherwise matched
STL-SFT settings, we compare randomly initialized STL encoders with
encoders initialized from semantic pre-training. All encoder parameters
remain trainable during policy fine-tuning. Semantic initialization raises  STL satisfaction from $46.7\%$ to
$61.7\%$ on seen specifications and from $53.0\%$ to $59.5\%$ on unseen
parameters. These results
show that robust semantics pre-training provides a more effective
initialization than random initialization and that its benefit extends
to formulas excluded from training.

\paragraph{Structured STL encoding vs.\ prompting}
We compare the structured syntax-graph encoder with a prompt baseline
that removes the STL encoder and appends a textual rendering of the
formula to the natural-language instruction. The prompt baseline
otherwise follows the same two-stage post-training procedure. Structured encoding provides its largest advantage on seen
specifications, increasing STL satisfaction from $66.5\%$ to $82.0\%$
while retaining comparable task success. It also improves
unseen-formula satisfaction from $70.1\%$ to $74.8\%$, with task
success increasing from $89.7\%$ to $92.2\%$. These results show that
direct prompting can communicate useful requirement information, while
explicitly encoding the formula's predicates, temporal operators, and
logical composition empirically provides a stronger conditioning signal.\vspace{-5pt}

\section{Conclusion}
We propose Logic-VLA, a temporal logic conditioned VLA. Logic-VLA combines a semantically pre-trained syntax-graph encoder with two-stage post-training using satisfying demonstrations and matched satisfying-violating trajectory pairs. In closed-loop quadcopter navigation, it substantially improves STL satisfaction while minimally reducing NL task success compared to an STL-blind base policy. The results show that formal logic can serve as effective conditioning signals for adapting a single VLA to varying requirements, including safety-critical and spatiotemporal specifications.\vspace{-5pt}

\ackswitch{\section{Acknowledgement}
We thank Ruohai Ge for helpful writing advice. The USC Physical Superintelligence Lab acknowledges the generous support from Toyota Research Institute, Dolby, Google DeepMind, Capital One, Nvidia, Bosch, NSF, and Qualcomm. This work was partially supported by the National Science Foundation through the following grants: CAREER award (SHF-2048094), 2434460, IIS-SLES-2417075, and funding by Toyota R\&D through the USC Center for Autonomy and AI. Yue Wang is supported by a Powell Research Award.}{}

\bibliographystyle{ieeetr}
\bibliography{main} 
\clearpage
\fullswitch{\section{Appendix}
\subsection{Semantics of Signal Temporal Logic}
\label{app:semantics_stl}
The semantics of an STL formula $\phi$ enabled at time $\tau$ is recursively defined as follows \cite{donze2010robust, fainekos2009robustness, zhao2024robust}, where $(s, \tau) \models \phi$ denote that $s$ satisfies $\phi$ at time $\tau$.
\begin{align*}
	(s,\tau)\models \text{True} & \hspace{0.1cm} \text{iff} \hspace{0.1cm} \text{True},\\
    (s, \tau) \models \pi^\mu & \hspace{0.1cm} \text{iff} \hspace{0.1cm} \mu(s_\tau) 
    \ge 0,\\
    (s, \tau) \models \neg \phi & \hspace{0.1cm} \text{iff} \hspace{0.1cm} (s, \tau) \not\models \phi,\\
    (s, \tau) \models \phi_1 \wedge \phi_2 & \hspace{0.1cm} \text{iff} \hspace{0.1cm} (s, \tau) \models \phi_1 \text{ and } (s, \tau) \models \phi_2, \\
    (s, \tau) \models \phi_1 U_{[a, b]} \phi_2 & \hspace{0.1cm} \text{iff} \hspace{0.1cm} \exists \tau_2 \in (\tau \oplus [a, b]) \cap \mathbb{N} \text{ s.t. }  (s, \tau_2) \models \phi_2 \\
    & \hspace{0cm} \text{ and } \hspace{0cm} \forall \tau_1 \in (\tau, \tau_2) \cap \mathbb{N}, (s, \tau_1) \models \phi_1.
\end{align*}
The robust semantics $\rho^\phi(s, \tau_0) \in \mathbb{R}$, measuring how robustly a formula is satisfied by a trajectory, is also recursively defined following the standards:
\begin{align*}
    \rho^\phi(s, \tau) &\coloneq \infty, \\
    \rho^{\pi^\mu}(s, \tau) &\coloneq \mu(s_\tau), \\
    \rho^{\neg\phi}(s, \tau) &\coloneq -\rho^\phi(s, \tau), \\
    \rho^{\phi_1\wedge\phi_2} &\coloneq \min\Big(\rho^{\phi_1}(s, \tau), \rho^{\phi_2}(s, \tau)\Big),\\
    \rho^{\phi_1 U_{[a, b]}\phi_2} &\coloneq \sup_{\tau_2 \in (\tau \oplus [a, b]) \cap \mathbb{N}}\Big(\min(\rho^{\phi_2}(s, \tau_2), \\ &\inf_{\tau_1\in (\tau, \tau_2) \cap \mathbb{N}}\rho^{\phi_1}(s, \tau_1)\Big).
\end{align*}
The formula length of $\phi$ (i.e. the termination time), $H^\phi \in \mathbb{N}$, denotes the sufficient length of the trajectory $(s_{\tau}, \hdots, s_{\tau+H})$ so that $(s, \tau) \models \phi$ can be evaluated \cite{sadraddini2015robust, zhao2024robust}.
\begin{align*}
    H^{\text{True}} &\coloneq 0,\\
    H^{\pi^\mu} &\coloneq 0, \\
    H^{\neg \phi} &\coloneq H^\phi, \\
    H^{\phi_1 \wedge \phi_2} &\coloneq \max(H^{\phi_1}, H^{\phi_2}),\\
    H^{\phi_1 U_{[a, b]} \phi_2} &\coloneq \max([a, b] \cap \mathbb{N}) + \max(H^{\phi_1}, H^{\phi_2}).
\end{align*}

\section{CRATE: Certified Randomized Trajectory Ensembles}
\label{app:crate}
We disclose the details for generating the dataset $D$ in   Section \ref{sec:eval}. Our dataset consists of smooth flight trajectories that are provably collision-free over continuous time, visit prescribed target regions in order, respect dynamic limits, and are diverse under controllable randomness. Compared to our data collection pipeline, sampling-based planners \cite{lavalle1998rapidly, kavraki1996probabilistic} produce varied but jagged paths with uncontrollable randomness and continuous-time safety re-established only after smoothing. On the other hand, corridor-based convex planners \cite{liu2017planning, gao2018online} are provably safe by construction but return a single optimal trajectory with no diversity interface. To our best knowledge, recent data-generating pipelines for learned planners \cite{mao2025rapid, takubo2026language} obtain diversity by randomizing the problem instances. We propose \textit{CRATE} (\textbf{C}ertified \textbf{RA}ndomized \textbf{T}rajectory \textbf{E}nsembles), an automatic trajectory data collection pipeline built upon the standard corridor and Bézier-QP architecture with randomization acting on the optimization objective. We release the codes in \href{https://celinawa.github.io/crate/}{https://celinawa.github.io/crate/}. We first formally state the problem instance we face.

\begin{problem}
\label{problem:data}
    A problem instance is given by axis-aligned boxes (AAB) and scalar limits: an obstacle-free start region $S$, from which each trajectory's start $s_0$ is sampled; ordered target AABs $G_1, \hdots, G_\xi$ with pairwise-distinct centers $w_k \in \mathbb{R}^3$ and half-widths $\rho_k \in \mathbb{R}^3_{\ge 0}$, i.e., $G_k = \{p : |p - w_k| \le \rho_k \text{ componentwise}\}$; obstacle AABs $O_1, 
    \hdots, O_\gamma$ defining free space $\mathcal{F} = \mathbb{R}^3 \setminus \bigcup_i \mathrm{int}(O_i)$; workspace walls $R$ and dynamics constraints $v_{\max}, a_{\max}, T_{\max}$. Given a problem instance, a trajectory $r : [0,T] \to \mathbb{R}^3$ is \textit{admissible} iff the following conditions hold: 1) $r(t) \in \mathcal{F} \cap R, \forall t \in [0, T]$; 2) $r(0) = s_0 \in S$; 3) $\exists \tau_1 < \dots < \tau_\xi = T$ with $r(\tau_k) \in G_k$; 4) per-axis $\lVert \dot r(t)\rVert_\infty \le v_{\max}$,
$\lVert\ddot r(t)\rVert_\infty\le a_{\max}, \forall t$ with $T \le T_{\max}$; 5) $r(t)$ departs and arrives at rest. Our task is to generate $K_i$ admissible trajectories per instance $i$ whose spread is controlled by a diversity parameter.
\end{problem}

Each NL task from Section~\ref{sec:eval} is abstracted into a formalized problem instance from Problem~\ref{problem:data} with the expert demonstrations $D$ generated using the following pipeline.

\subsection{Method}
CRATE first commits, deterministically, to where the trajectories go. Let $O_i^+ = O_i \oplus [-m, m]^3$ be the obstacles inflated by a margin $m$, with $m > \delta$ (the corridor clearance below), so a planned route always leaves room for a $\delta$-clear corridor. Per instance, CRATE builds a roadmap $\mathcal{R} = (V, E, \ell)$ once, with vertices $V$ the start-region center, the target centers $w_1, \hdots, w_\xi$, and a workspace grid clear of $\bigcup_i O_i^+$, and edges $E = \{(u,v) : \|u-v\| \le r_c,\ \overline{uv} \cap O_i^+ = \emptyset\ \forall i\}$ weighted by length $\ell(u,v) = \|u-v\|$. A \textit{route} is a shortest polyline over $\mathcal{R}$ through $w_1, \hdots, w_\xi$ in order; up to $n_r$ distinct routes $\pi_1, \hdots, \pi_{n_r}$ are extracted by multiplying the weights of each found route's edges by a penalty $\beta > 1$ before the next query, a standard device for diverse alternative paths \cite{yen1971finding, bhattacharya2012topological}. This layer only fixes each route's homotopy and carries no safety burden. Each $\pi_j$ is then inflated into a \textit{safe flight corridor} \cite{liu2017planning} $\mathcal{B}_j = (B_1, \hdots, B_{M_j})$: boxes grown face by face from the segments of $\pi_j$ up to a clearance $\delta$ from all obstacles, satisfying (C1) $B_q \cap O_i = \emptyset\ \forall q, i$; (C2) $B_q \cap B_{q+1} \neq \emptyset$; (C3) $\pi_j \subset \bigcup_q B_q$. If the instance has a start region, $B_1$ is grown from $S$ itself, so $S \subseteq B_1$. A route admitting no corridor satisfying (C1)--(C3) is discarded --- CRATE never emits an unsafe tube. Within each corridor, a convex program generates the trajectories, as described next.

Within a corridor $\mathcal{B}_j$, each trajectory is a spline of $M_j$ Bézier segments, one per box: segment $q$ has degree $d$, control points $c_{q,0}, \hdots, c_{q,d} \in \mathbb{R}^3$, and duration $\Delta_q = T\, w_q / \sum_p w_p$, splitting a total time $T \le T_{\max}$ across the boxes according
to a positive weight vector $w \in \mathbb{R}^{M_j}_{>0}$, so $\sum_q \Delta_q = T$ for any $w$; by default $w_q = \ell
_q$, the length of the $q$-th segment of $\pi_j$. A Bézier curve lies, at every parameter value, in the convex hull of its control points, and hence in any convex set containing them; its derivatives are again Bézier curves whose control points are linear in the $c_{q,i}$ \cite{farouki2012bernstein}. Writing $r_q : [0, \Delta_q] \to \mathbb{R}^3$ for segment $q$'s Bézier curve so that the trajectory $r$ traverses $r_1, \hdots, r_{M_j}$ over consecutive intervals of lengths $\Delta_q$, each trajectory within $\mathcal{B}_j$ is obtained by solving jointly for all control points, confining each segment's control points to its box as in \cite{gao2018online}:

\begin{subequations}
\label{eq:qp}
\begin{align}
\min_{c} \;\; & \textstyle\sum_q \int_0^{\Delta_q} \lVert r_q^{(4)}(t) \rVert^2 dt \,+\, \lambda \sum_q \lVert r_q(\Delta_q/2) - \alpha_q \rVert^2 \label{eq:qp:obj} \\
\text{s.t.} \;\; & c_{q,i} \in B_q \cap R \;\; \forall q, i; \qquad c_{1,0} = s_0; \label{eq:qp:safe} \\
& \dot r(0) = \ddot r(0) = \dot r(T) = \ddot r(T) = 0; \label{eq:qp:rest} \\
& r_q^{(o)}(\Delta_q) = r_{q+1}^{(o)}(0), \;\; o \le 3, \; q < M_j; \label{eq:qp:cont} \\
& c_{q_k,d} \in \widetilde G_k, \;\; k = 1, \hdots, \xi; \label{eq:qp:reach} \\
& c^{(1)}_{q,i} \in [-v_{\max}, v_{\max}]^3, \;\; c^{(2)}_{q,i} \in [-a_{\max}, a_{\max}]^3 \;\; \forall q, i, \label{eq:qp:dyn}
\end{align}
\end{subequations}
where $c$ collects all control points; $q_k$ is the segment of $\mathcal{B}_j$ at whose end boundary the route visits $G_k$, so the end point $c_{q_k,d}$ lies on the curve; $\widetilde G_k$ is the concentric inner box of $G_k$, with half-widths halved on axes wider than $0.2$\,m so that arrivals register as visits rather than boundary grazes; $\alpha_q = \mathrm{clip}_{B_q}(b_q + \sigma h_q \odot \eta_q)$, $\eta_q \sim \mathcal{N}(0, I)$, is a random attractor with $b_q, h_q$ the center and half-widths of $B_q$; and $c^{(1)}_{q,i} = \tfrac{d}{\Delta_q}\big(c_{q,i+1} - c_{q,i}\big)$, $i = 0, \hdots, d-1$, and $c^{(2)}_{q,i} = \tfrac{d(d-1)}{\Delta_q^2}\big(c_{q,i+2} - 2c_{q,i+1} + c_{q,i}\big)$, $i = 0, \hdots, d-2$, are the control points of $\dot r_q$ and $\ddot r_q$, linear in $c$, so \eqref{eq:qp:dyn} bounds $\dot r$ and $\ddot r$ everywhere by the convex-hull property. In general, $c^{(o)}_{q,i} = \tfrac{d!}{(d-o)!\,\Delta_q^o}(\nabla^o c_q)_i$, $i = 0, \hdots, d-o$, with $\nabla^o$ the $o$-fold forward difference; since a Bézier curve interpolates its first and last control points, $r_q^{(o)}(\Delta_q) = c^{(o)}_{q,d-o}$ and $r_{q+1}^{(o)}(0) = c^{(o)}_{q+1,0}$, each instance of \eqref{eq:qp:cont} is a single linear equality between the two segments' control points, with each side scaled by its own $\Delta^{-o}$. The first objective term is the minimum-snap cost \cite{mellinger2011minimum}; the second pulls each segment midpoint toward its attractor. Substituting $r_q(t) = \sum_{i=0}^{d} \beta_{i,d}(t/\Delta_q)\, c_{q,i}$, where $\beta_{i,d}$ is the Bernstein basis, the snap term becomes $\sum_q c_q^\top Q_q c_q$ with $Q_q = \Delta_q^{-7} D^\top \bar G D \succeq 0$ for constant matrices $D$ (the fourth-difference operator scaled by $d!/(d-4)!$) and $\bar G$ (the Bernstein Gram matrix of the degree-$(d-4)$ Bernstein basis), and the midpoint $r_q(\Delta_q/2) = \beta^\top_{1/2} c_q$ is linear in $c_q$ so that the objective of \eqref{eq:qp} is a convex quadratic and every constraint is linear in $c$. Each of the $K_i$ draws samples fresh attractors, a total time $T$ uniform in a range $[T_{\mathrm{lo}}, T_{\mathrm{hi}}] \subseteq (0, T_{\max}]$, optionally duration weights $w \sim \mathrm{Dir}(\kappa \bar\ell)$ with $\bar\ell_q = \ell_q / \sum_p \ell_p$, and a start $s_0$ uniform over $S$. Randomization touches only the objective, the boundary value $s_0$ within $S \subseteq B_1$, and the time allocation $\{\Delta_q\}$: the safety and reach constraints \eqref{eq:qp:safe}, \eqref{eq:qp:reach} are identical across draws, and each draw's dynamics constraints are enforced at its own timing. Problem \eqref{eq:qp} decouples per axis into three QPs, solved with OSQP \cite{stellato2020osqp}; when all draws share one timing, the KKT factorization is assembled once per route and reused across the whole batch. Draws for which \eqref{eq:qp} is infeasible (an overly aggressive $T$) are dropped. We show the derivations of snap-matrix reduction in Section~\ref{subsec:snap_matrix}.

\begin{proposition}
\label{prop:admissible}
Every trajectory emitted by CRATE is admissible for its problem instance in the sense of Problem~\ref{problem:data}.
\begin{proof}
    Fix a surviving route $\pi_j$ with corridor $\mathcal{B}_j$ satisfying (C1)--(C3) and $S \subseteq B_1$, and a draw $(\alpha, T, w, s_0)$ with $T \in [T_{\mathrm{lo}}, T_{\mathrm{hi}}] \subseteq (0, T_{\max}]$ and $s_0 \in S$ for which \eqref{eq:qp} is feasible; let $c$ be a feasible point and $r$ the resulting spline. Since $T > 0$ and $w > 0$, every $\Delta_q > 0$, so the knot times $t_q = \sum_{p \le q} \Delta_p$ are strictly increasing with $t_{M_j} = T$. We verify conditions 1)--5) of Problem~\ref{prob:overall} in turn; throughout we use that a Bézier curve lies in the convex hull of its control points, interpolates its first and last, and has derivative control points $c^{(o)}_{q,i}$ as given above \cite{farouki2012bernstein}. \emph{1) Free space and walls.} Any $t \in [0,T]$ lies in some segment interval, where $r(t) = r_q(t - t_{q-1})$. By \eqref{eq:qp:safe} each $c_{q,i} \in B_q \cap R$, a convex set, so $r(t) \in \mathrm{conv}\{c_{q,i}\}_i \subseteq B_q \cap R$. By (C1), $B_q \cap O_i = \emptyset$ for every obstacle, hence $B_q \subseteq \mathcal{F}$ and $r(t) \in \mathcal{F} \cap R$ for all $t$, with no time sampling. \emph{2) Start.} $r(0) = r_1(0) = c_{1,0} = s_0 \in S$ by interpolation and \eqref{eq:qp:safe}. \emph{3) Ordered reach.} The route visits $w_1, \hdots, w_\xi$ in the prescribed order, and each additional route segment contributes at least one box, so the boundary indices are strictly increasing: $q_1 < q_2 < \hdots < q_\xi = M_j$ (the final target is the route's last vertex). Set $\tau_k = t_{q_k}$; then $\tau_1 < \hdots < \tau_\xi = T$, and by interpolation and \eqref{eq:qp:reach}, $r(\tau_k) = r_{q_k}(\Delta_{q_k}) = c_{q_k,d} \in \widetilde G_k \subseteq G_k$. \emph{4) Dynamics.} On the interior of segment $q$, $\dot r(t) = \sum_i \beta_{i,d-1}\!\big((t - t_{q-1})/\Delta_q\big)\, c^{(1)}_{q,i} \in \mathrm{conv}\{c^{(1)}_{q,i}\}_i \subseteq [-v_{\max}, v_{\max}]^3$ by \eqref{eq:qp:dyn}, and likewise $\ddot r(t) \in [-a_{\max}, a_{\max}]^3$. At the knots, \eqref{eq:qp:cont} matches derivatives up to order $3$, so $\dot r$ and $\ddot r$ are well defined there and the bounds extend to all $t \in [0,T]$. Finally $T \le T_{\max}$ since the drawn range lies in $(0, T_{\max}]$. \emph{5) Rest.} Immediate from \eqref{eq:qp:rest}. The argument used only the feasibility of $c$ in \eqref{eq:qp} --- never optimality, nor the values of $\alpha$ or $w$ --- and randomization alters neither \eqref{eq:qp:safe} nor \eqref{eq:qp:reach}: Every draw with a solved QP yields an admissible trajectory.
\end{proof}
\end{proposition}

\subsection{Snap-matrix reduction}
\label{subsec:snap_matrix}
The factored form of $Q_q$ is standard \cite{mellinger2011minimum, farouki2012bernstein}; we re-derive it for clarity. Fix one axis and write $x_q \in \mathbb{R}^{d+1}$ for segment $q$'s control-point coordinates on it, so $r_q(t) = \sum_i \beta_{i,d}(u)\, x_{q,i}$ in scaled time $u = t/\Delta_q$. Differentiating a Bézier curve in $u$ lowers its degree by one and maps its control points by a degree-scaled forward difference; four rounds give $\hat r_q^{(4)}(u) = \sum_{i=0}^{d-4} \beta_{i,d-4}(u) (Dx_q)_i$, writing $\hat r_q(u) = r_q(\Delta_q u)$ for the curve in scaled time, with $D = \tfrac{d!}{(d-4)!}\nabla^4$, where $\nabla^4$ is the four-fold forward-difference matrix. Four chain rules $\tfrac{d}{dt} = \Delta_q^{-1}\tfrac{d}{du}$ and the substitution $dt = \Delta_q du$ then yield
\begin{equation}
\begin{split}
\int_0^{\Delta_q} \big(r_q^{(4)}\big)^2 dt \;&=\; \Delta_q^{-8}\cdot\Delta_q \int_0^1 \big(\hat r_q^{(4)}\big)^2 du \\
&=\; \Delta_q^{-7} (Dx_q)^\top \bar G\, (Dx_q) \;=\; x_q^\top Q_q x_q,
\end{split}
\label{eq:snapmatrix}
\end{equation}
where $\bar G_{ij} = \int_0^1 \beta_{i,n}(u)\,\beta_{j,n}(u)\, du = \binom{n}{i}\binom{n}{j}/\big(\binom{2n}{i+j}(2n+1)\big)$, $n = d-4$, in closed form since a product of two Bernstein polynomials is a rescaled Bernstein polynomial of degree $2n$, each integrating to $\tfrac{1}{2n+1}$ \cite{farouki2012bernstein}. As a congruence transform of a Gram matrix, $Q_q \succeq 0$; $D$ and $\bar G$ depend only on $d$, so across all segments only the scalar $\Delta_q^{-7}$ varies.

\subsection{Dataset}
We provide here details of the parameters used for generating the dataset $D$ in Section \ref{sec:eval}. We instantiate Problem~\ref{problem:data} in 10 warehouse environments with the tasks described in Section \ref{sec:eval} with obstacle geometry from randomized warehouse assets and target boxes placed near off asset surfaces, sized for a DJI Mavic 2 Pro. All instances share one frozen configuration: $n_r = 5$, $\beta = 20$, $30$ draws per route, $\sigma = 0.1$, $\lambda = 8$, $[T_{\mathrm{lo}}, T_{\mathrm{hi}}] = [7, 15]$\,s, $d = 5$, $\delta = 0.27$\,m, $m = 0.32$\,m, $v_{\max} = 5$\,m/s, $a_{\max} = 10$\,m/s$^2$, $T_{\max} = 30$\,s. The dataset $D$ contains up to $50$ trajectories per task as $10$\,Hz waypoints with positions, velocities, and accelerations; each trajectory was additionally re-validated against all five conditions of Problem~\ref{problem:data} at the delivered waypoint resolution.

\section{Data Curation}
\label{sec:stl_generator}

\paragraph{Comparable rollout groups} Consider $s^{(i)}$, the trajectory component of $p^{(i)} \in D$. To ensure that STL satisfaction and violation are compared across executions of the same nominal behavior, we construct a finite collection of comparable rollout groups $\mathcal{C} \coloneq \{C_1, \hdots, C_{N_C}\}$ where $C_k \subseteq D$. Each $C_k$ contains rollouts that share the same nominal NL  task and environment and satisfy a prescribed notion of initial-condition comparability. The benchmark specific realization of the grouping procedure is shown in Appendix \ref{sec:stl_generator}. Let $\mathcal{Q}_0 \coloneq \mathcal{C} \times \Phi$ define the set of candidate group-formula combinations. For each $(C, \phi) \in \mathcal{Q}_0$ and each sample $p^{(i)} \in C$, an offline STL monitor computes $\rho^{\phi, (i)} \coloneq \rho^\phi(s^{(i)})$.

\paragraph{Informative formula-group selection} Not every element of $\mathcal{Q}_0$ provides meaningful supervision. We define a selection rule $\mathcal{F}:\mathcal{Q}_0 \rightarrow \{0, 1\}$ and retain $\mathcal{Q} \coloneq \{(C, \phi) \in \mathcal{Q}_0 \mid \mathcal{F}(C, \phi) = 1\}$. Conceptually, $\mathcal{F}$ retains group-formula combinations that provide suitable positive and negative examples. For each retained $(C, \phi) \in \mathcal{Q}$, given robust semantics margins $\epsilon_+, \epsilon_- > 0$, we define the clearly satisfying and violating subsets $C^+_\phi \coloneq \{p^{(i)} \in C \mid \rho^{\phi, (i)} \ge \epsilon_+\}$ and $C^-_\phi \coloneq \{p^{(i)} \in C \mid \rho^{\phi, (i)} \le -\epsilon_-\}$. In our instantiation, $\mathcal{F}$ requires $C^+_\phi ,C^-_\phi \neq \emptyset$, together with additional criteria ensuring that the induced satisfying-violating distinction is informative. For each retained $(C, \phi)$, let $\tilde{C}^+_\phi \subseteq C_\phi^+$ denote the satisfying rollouts retained for SFT. We next define a trajectory-pair selection rule $\mathcal{G}_{C, \phi}:C_\phi^+\times C_\phi^- \rightarrow \{0, 1\}$ where $\mathcal{G}_{C, \phi}(p^+, p^-) = 1$ indicates that the satisfying rollout $p^+$ and violating rollout $p^-$ form an admissible matched counterexample. The final set of selected preference pairs is $\tilde{M}_{C, \phi} \coloneq \{(p^+, p^-) \in C_\phi^+ \times C_\phi^- \mid \mathcal{G}_{C, \phi}(p^+, p^-) = 1\}$. We require $\tilde{M}_{C, \phi} \neq \emptyset$ for every retained $(C, \phi)$, so that every retained combination contributes preference supervision. The exact realization of $\mathcal{F}$ and $\mathcal{G}_{C, \phi}$ is benchmark-specific and is provided in Appendix \ref{sec:stl_generator}.

\paragraph{Training set construction} The selected satisfying rollouts and matched counterexamples define the two post-training datasets. For each $p^{(i)} \in \tilde{C}^+_\phi$, at time $\tau$, let $\xi_{i,\tau}=(o^{(i)}_\tau,l^{(i)},\phi)$ be the STL-conditioned context and $a^{(i)}_\tau$ be the demonstrated action chunk. We construct the positive demonstration dataset $D^+ \coloneq \{(\xi_{i,\tau}, a^{(i)}_\tau) \mid (C, \phi) \in \mathcal{Q}, p^{(i)} \in \tilde{C}^+_\phi, \tau \in W_i\}$, where $W_i$ denotes the set of temporal windows from which action chunks are extracted. The selected matched counterexamples directly define the trajectory-level preference dataset $P \coloneq \{(p^+, p^-, l, \phi) \mid (C, \phi) \in \mathcal{Q}, (p^+, p^-) \in \tilde{M}_{C, \phi}\}$, where $l$ is the shared nominal NL task. The resulting supervision consists of satisfying demonstrations $D^+$ for STL-conditioned SFT and matched satisfying-violating trajectory pairs $P$ for trajectory-level preference optimization.

\section{Architectural and Experiment Details}
\label{app:arch}
\paragraph{Conditioning location}

We compare action-expert-side and VLM-side conditioning under the same
two-stage post-training procedure.

\begin{table}[t]
    \centering
    \caption{
        Effect of specification-token placement under the complete
        two-stage post-training procedure. Values are percentages.
        Bold indicates the numerically higher value within each split.
    }
    \label{tab:conditioning_location_ablation}
    \scriptsize
    \setlength{\tabcolsep}{3.0pt}
    \renewcommand{\arraystretch}{1.05}
    \begin{tabular}{lcccc}
        \hline
        &
        \multicolumn{2}{c}{Seen}
        &
        \multicolumn{2}{c}{Unseen Formula} \\
        \cline{2-5}
        Conditioning
        & STL Sat. $\uparrow$
        & Task $\uparrow$
        & STL Sat. $\uparrow$
        & Task $\uparrow$ \\
        \hline

        Action expert
        & 74.5
        & \textbf{92.7}
        & \textbf{76.1}
        & 91.6 \\

        Vision--language
        & \textbf{82.0}
        & 89.0
        & 74.8
        & \textbf{92.2} \\

        \hline
    \end{tabular}
\end{table}

On seen specifications, VLM-side conditioning achieves higher STL
satisfaction ($82.0\%$ vs.\ $74.5\%$), while action-expert-side
conditioning retains higher task success ($92.7\%$ vs.\ $89.0\%$).
On unseen formulas, the two configurations yield similar point
estimates, differing by $1.3$ percentage points in STL satisfaction and
$0.6$ points in task success. The same pattern holds on unseen
structures, where their satisfaction and task-success rates differ by
only $1.0$ and $0.2$ points, respectively. Thus, neither conditioning
location is uniformly superior, and both support comparable
generalization to unseen specifications. Full unseen-structure results
are provided in the Appendix.

\begin{table}[t]
    \centering
    \scriptsize
    \setlength{\tabcolsep}{3.5pt}
    \renewcommand{\arraystretch}{1.08}

    \begin{tabular}{lcc|cc|cc}
        \hline
        &
        \multicolumn{2}{c|}{Seen}
        &
        \multicolumn{2}{c|}{Unseen Param.}
        &
        \multicolumn{2}{c}{Unseen Struct.} \\
        \cline{2-7}

        Method
        & Coll. $\downarrow$
        & Safe $\uparrow$
        & Coll. $\downarrow$
        & Safe $\uparrow$
        & Coll. $\downarrow$
        & Safe $\uparrow$ \\
        \hline

        Base
        & \textbf{15.8}
        & \textbf{78.8}
        & \textbf{10.4}
        & \textbf{86.2}
        & \textbf{16.1}
        & \textbf{78.7}
        \\

        STL-SFT
        & 17.0
        & 73.8
        & 12.9
        & 79.6
        & 18.2
        & 71.1
        \\

        Smooth ($1\times$)
        & 43.5
        & 44.2
        & 36.5
        & 53.5
        & 43.2
        & 45.2
        \\

        Smooth ($2\times$)
        & 47.0
        & 31.8
        & 45.4
        & 29.2
        & 47.9
        & 24.6
        \\

        \rowcolor{blue!8}
        \textbf{Logic-VLA}
        & 20.7
        & 72.7
        & 13.8
        & 79.8
        & 20.7
        & 72.1
        \\

        \hline
    \end{tabular}

    \caption{\textit{[Caption to be added.]}}
    \label{tab:safety_results}
\end{table}

\begin{table}[t]
    \centering
    \caption{
        Effect of semantic STL-encoder initialization under the default
        VLM-side STL-SFT configuration. Values are percentages.
    }
    \label{tab:semantic_pre-training_ablation}
    \scriptsize
    \setlength{\tabcolsep}{2.8pt}
    \renewcommand{\arraystretch}{1.05}
    \begin{tabular}{lcccc}
        \hline
        &
        \multicolumn{2}{c}{Seen}
        &
        \multicolumn{2}{c}{Unseen Formula} \\
        \cline{2-5}
        Initialization
        & STL Sat. $\uparrow$
        & Task $\uparrow$
        & STL Sat. $\uparrow$
        & Task $\uparrow$ \\
        \hline
        Random
        & 46.7
        & 84.5
        & 53.0
        & 89.4 \\

        Semantic pre-training
        & \textbf{61.7}
        & \textbf{87.2}
        & \textbf{59.5}
        & \textbf{91.1} \\
        \hline
    \end{tabular}
\end{table}
\begin{table}[t]
    \centering
    \caption{
        Comparison of direct textual prompting and structured STL
        encoding on seen and unseen-formula specifications under the
        complete two-stage post-training procedure. Values are
        percentages. Bold indicates the numerically higher value within
        each split.
    }
    \label{tab:representation_ablation}
    \scriptsize
    \setlength{\tabcolsep}{2.8pt}
    \renewcommand{\arraystretch}{1.05}
    \begin{tabular}{lcccc}
        \hline
        &
        \multicolumn{2}{c}{Text prompt}
        &
        \multicolumn{2}{c}{Structured encoder} \\
        \cline{2-5}
        Split
        & STL Sat. $\uparrow$
        & Task $\uparrow$
        & STL Sat. $\uparrow$
        & Task $\uparrow$ \\
        \hline

        Seen
        & 66.5
        & 88.5
        & \textbf{82.0}
        & \textbf{89.0} \\

        Unseen Formula
        & 70.1
        & 89.7
        & \textbf{74.8}
        & \textbf{92.2} \\

        \hline
    \end{tabular}
\end{table}}{}
\end{document}